%% file: main.tex
\documentclass{article}
\usepackage[final]{neurips_2025}
\pdfoutput=1
\PassOptionsToPackage{numbers}{natbib}

\usepackage[utf8]{inputenc} 
\usepackage[T1]{fontenc}    
\usepackage{hyperref}       
\usepackage{url}            
\usepackage{booktabs}       
\usepackage{amsfonts}       
\usepackage{nicefrac}       
\usepackage{microtype}      
\usepackage{xcolor}         
\usepackage{amsmath}
\usepackage{enumitem}
\usepackage{graphicx}
\usepackage{wrapfig}

\usepackage{multirow}
\usepackage{xcolor}
\usepackage{colortbl}   
\usepackage{tabularx}
\usepackage{float}
\usepackage{tcolorbox}
\usepackage{makecell}

\usepackage{pifont}

\definecolor{deepgreen}{RGB}{0, 100, 0}
\definecolor{deepred}{RGB}{139, 0, 0}

\definecolor{backblue}{RGB}{210, 230, 250}

\usepackage{xspace}

\definecolor{backblue}{RGB}{210, 230, 250}
\definecolor{backgreen}{RGB}{226, 240, 217}
\definecolor{backred}{RGB}{255, 223, 223}

\definecolor{citecolor}{RGB}{0,255,0}
\definecolor{refcolor}{RGB}{255,0,0}
\definecolor{urlcolor}{RGB}{0,180,255}

\hypersetup{
    colorlinks=true,
    linkcolor=refcolor,
    citecolor=citecolor,
    urlcolor=urlcolor,
}

\title{Look Before You Leap: A GUI-Critic-R1 Model for Pre-Operative Error Diagnosis in GUI Automation}

\author{Yuyang Wanyan$^{1,2,3}$\footnotemark[2] \footnotemark[1] ~, Xi Zhang$^{3}$\footnotemark[1] , Haiyang Xu$^{3}$\footnotemark[3] , Haowei Liu$^{3}$, Junyang Wang$^{3}$, Jiabo Ye$^{3}$,\\ 
\textbf{Yutong Kou$^{1}$, Ming Yan$^{3}$\footnotemark[3] , Fei Huang$^{3}$, Xiaoshan Yang$^{1,2}$\footnotemark[3] , Weiming Dong$^{1,2}$, Changsheng Xu$^{1,2}$ }\\
$^{1}$MAIS, Institute of Automation, Chinese Academy of Sciences, China \\
$^{2}$School of Artificial Intelligence, University of Chinese Academy of Sciences, China \\
$^{3}$Alibaba Group \\
wanyanyuyang2021@ia.ac,cn,  xiaoshan.yang@nlpr.ia.ac.cn, \\
\{shuofeng.xhy, ym119608\}@alibaba-inc.com\\
}

\begin{document}

\maketitle


\renewcommand{\thefootnote}{\fnsymbol{footnote}}

\footnotetext[1]{Equal contribution.~ 
$^{\dag}$Work done during internship at Tongyi Lab, Alibaba Group.~ 
$^{\ddag}$Corresponding Author.}


\input{sec/0_abstract}
\input{sec/1_intro}

\input{sec/2_related_work}

\input{sec/4_method}
\input{sec/5_experiment}

\input{sec/6_conclusion}

\section*{Acknowledgements}

This work was supported by the National Natural Science Foundation of China (Grants 62322212, U23A20387), the Beijing Natural Science Foundation (No. L221013), and the CAS Project for Young Scientists in Basic Research (YSBR-116).

\bibliographystyle{plain}
\bibliography{ref}

\input{sec/checklist}
\input{sec/appendix}

\end{document}

%% file: sec/0_abstract.tex
\begin{abstract}
In recent years, Multimodal Large Language Models (MLLMs) have been extensively utilized for multimodal reasoning tasks, including Graphical User Interface (GUI) automation. 
Unlike general offline multimodal tasks, GUI automation is executed in online interactive environments,
necessitating step-by-step decision-making based on the real-time status of the environment.
This task has a lower tolerance for decision-making errors at each step, as any mistakes may cumulatively disrupt the process and potentially lead to irreversible outcomes like \textit{deletions} or \textit{payments}. 
To address these issues, we introduce a \textbf{pre-operative critic} mechanism that provides effective feedback prior to the actual execution,
by reasoning about the potential outcome and correctness of actions. 
Specifically, we propose a Suggestion-aware Group Relative Policy Optimization (\textbf{S-GRPO}) strategy to construct our pre-operative critic model GUI-Critic-R1,
incorporating a novel suggestion reward to enhance the reliability of the model's feedback.
Furthermore, we develop a reasoning-bootstrapping based data collection pipeline to create a \textbf{GUI-Critic-Train} and a \textbf{GUI-Critic-Test}, filling existing gaps in GUI critic data.
Static experiments on the GUI-Critic-Test across both mobile and web domains reveal that our GUI-Critic-R1 offers significant advantages in critic accuracy compared to current MLLMs.
Dynamic evaluation on GUI automation benchmark further highlights the effectiveness and superiority of our model, as evidenced by improved success rates and operational efficiency.
The code is available at \url{https://github.com/X-PLUG/MobileAgent/tree/main/GUI-Critic-R1}.
\end{abstract}

%% file: sec/1_intro.tex
\section{Introduction}
\label{sec:intro}

Recently, Multi-modal Large Language Models (MLLMs), leveraging their remarkable perception and reasoning capabilities, have demonstrated outstanding capabilities in various domains~\cite{bai2025qwen2, wu2024deepseekvl}.
Among these,
GUI automation,
as a practical multi-modal application scenario, is emerging as a significant technological revolution in artificial intelligence interactions~\cite{wang2025mobile,liu2025pc,zhang2023appagent,li2024appagent,xu2024aguvis,qin2025ui,rawles2024androidworld}.
To be specific, given an online GUI device and a natural language instruction,
it requires the GUI agent driven by MLLMs
to generate
a series of precise operations similar to how humans do~\cite{zhang2024large,tang2025survey}, such as click, type, and scroll.

\begin{figure}[t]
  \centering
  \includegraphics[width=1.0\textwidth]{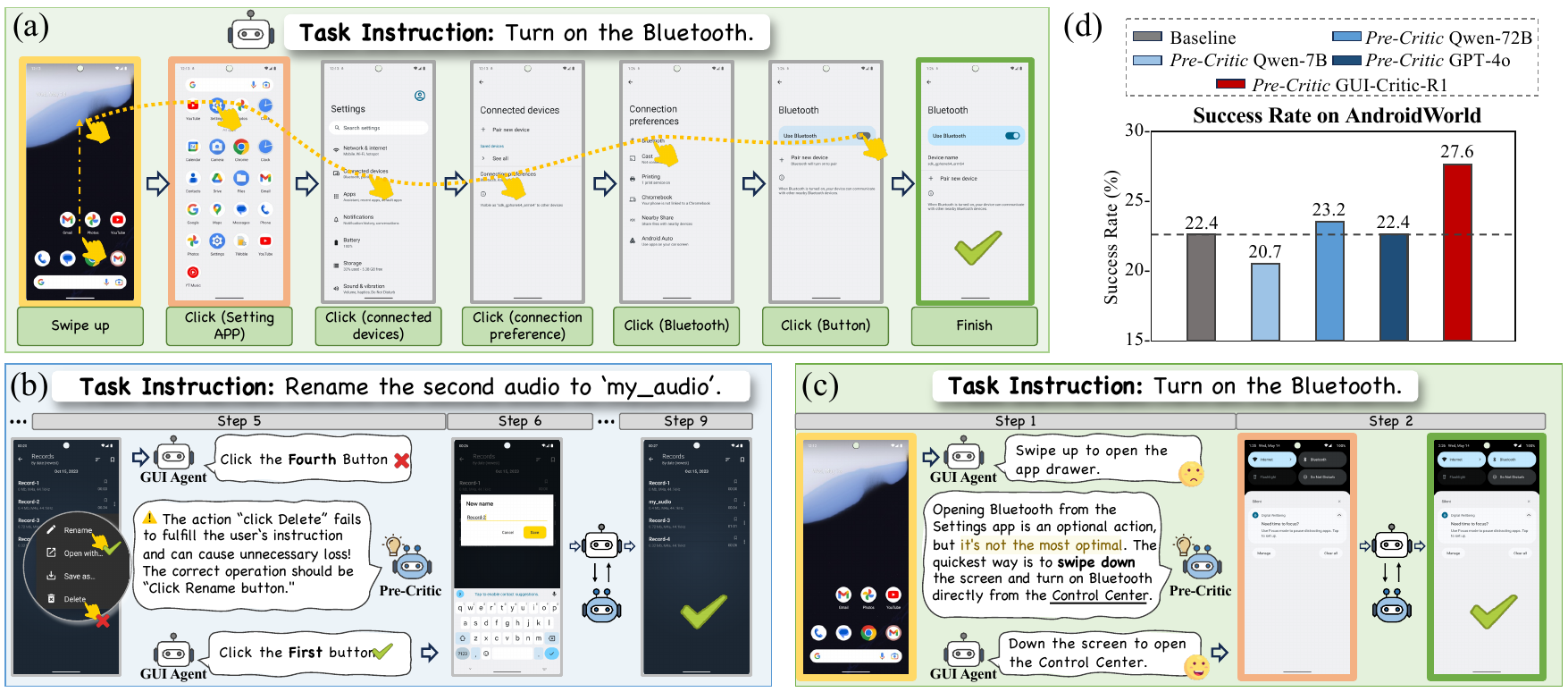}
  \vspace{-2mm}
  \caption{
(a) shows an example of GUI automation. 
 Case studies in (b-c) demonstrate how pre-critic prevents erroneous and redundant actions in GUI automation. 
(d) illustrates the quantitative performance comparison between pre-critic methods and baseline on AndroidWorld~\cite{rawles2024androidworld}.
}
\label{fig:overview}
  \vspace{-2mm}
\end{figure}

Unlike traditional offline multimodal tasks such as visual question answering~\cite{vqa_survey} and optical character recognition~\cite{ocr_survey}, GUI automation task operates within an online interactive environment and has some inherent key challenges~\cite{tang2025survey}. 
To be specific, the agents are required to generate coherent and sequential operations step by step.
In this way, 
an error in one step
can have cumulative effects on subsequent operations
(\textit{e.g., delete a file shown in Figure~\ref{fig:overview}(b)}), thereby disrupting the entire interactive process. 
However, 
constrained by limited reflection capabilities, 
current MLLMs usually struggle to detect errors independently~\cite{tyen2023llms}.
Therefore, to ensure the single-step accuracy, 
it is necessary to provide MLLM-based agents with additional feedbacks that incorporates critic analysis,
such as assessments of the action correctness, potential outcomes, and action suggestions.
Nevertheless, in the dynamic environment, erroneous operations usually require additional steps to correct (\textit{e.g., refunding after an incorrect payment}), and some dangerous errors may be irreparable (\textit{e.g., deleting files as shown in Figure~\ref{fig:overview}(b)}).
Therefore, to prevent these issues from happening,
we believe that the critical feedback should be provided to the agent \textbf{before the action is actually executed}.
Additionally, completing a user instruction on a GUI device typically entails multiple pathways.
From an application perspective, the GUI agent is expected to complete instructions with an optimal path that contains the least number of steps.
In this way, the prior critical feedback
may also prevent the model from selecting a sub-optimal path with more steps as shown in Figure~\ref{fig:overview}(a), thereby enhancing the efficiency of completing instructions.

Considering the above issues, in this paper, we propose a \textbf{pre-operative critic} (pre-critic) mechanism for GUI automation.
Specifically, 
before performing an operation in online environments,
the pre-critic mechanism first evaluates whether the operation generated by the agent is beneficial to the instruction completion, through comprehending the screenshot and analyzing potential results of the operation.
Then, it provides real-time critical feedback to the agent,
including the underlying causes of errors and corrective suggestions,
to assist in refining its decision-making process.
For example,
as shown in Figure~\ref{fig:overview}(b),
given the instruction \textit{Rename the current audio to `my\_audio'},
the agent initially predicts the action of \textit{click (delete button)} in the 5-th step.
Such operation may cause the audio file to be deleted and the task to fail.
Fortunately, with the pre-critic mechanism, we can catch the dangerous error before executing, and provide objective feedback to the agent.
Besides, Figure~\ref{fig:overview}(c) illustrates a case where the pre-critic identifies a more efficient method to \textit{turn on Bluetooth} (\textit{i.e., enabling Bluetooth via the control center}),
thereby reducing the operation steps.
We also exhibit some statistic results in Figure~\ref{fig:overview}(d),
where the pre-critic helps increase the success rate of baseline from 22.4\% to 27.6\% in the dynamic GUI automation~\cite{rawles2024androidworld}~\footnote{More detailed experimental results about efficiency can be found in Table~\ref{tab: android world}}.
In conclusion, these figures demonstrate that the introduction of pre-critic can effectively alleviate the aforementioned issues of error revision and operational inefficiency in online environments.

An intuitive approach to implementing a pre-critic model is leveraging existing MLLMs.
However, 
closed-source models~\cite{gpt4o} incur heavy efficiency and cost issues,
making them unsuitable for real-time GUI automation. 
Besides,
open-source models~\cite{chen2024expanding,bai2025qwen2,wu2024deepseekvl} struggle as pre-critic models due to inherent limitations in comprehending GUI interface and forecasting interaction outcomes~\cite{liu2025pc,wang2024mobile,zhang2024critic}.
To this end,
we propose a specialized 7B model \textbf{GUI-Critic-R1} that harmonizes performance and efficiency for GUI pre-critic.
In particular,
to enhance the model's GUI reasoning and generalization capabilities,
we introduce a Suggestion-aware Group Relative Policy Optimization (\textbf{S-GRPO}).
In S-GRPO, a suggestion reward is innovatively designed to refine the model's critic reasoning process and ensure it provides reliable suggestions for fixing the error operation.
Moreover,
since it lacks both training and test datasets for pre-critic in GUI automation,
we develop a reasoning-bootstrapping based data collection pipeline to construct a \textbf{GUI-Critic-Train} and a \textbf{GUI-Critic-Test} dataset.
Specifically, GUI-Critic-Train contains 6K high-quality chain-of-thought data about mobile devices for robust training.
GUI-Critic-Test includes 1k samples encompassing both mobile and web scenarios, and aims to explicitly evaluate the pre-critic model's diagnostic capabilities when exposed to novel instructions or applications.
In experiments, 
we first compare our GUI-Critic-R1 with advanced MLLMs on the GUI-Critic-Test,
and find that our model achieves satisfactory results in the critic accuracy.
Then, we also apply the GUI-Critic-R1 to a real-time GUI automation benchmark (\textit{i.e.}, AndroidWorld~\cite{rawles2024androidworld}),
 the improvements on the success rate further demonstrate the effectiveness of the proposed method. 

Our main contributions are summarized as follows:
\begin{enumerate}
    \item {
    To the best of our knowledge, we are the first to investigate a pre-operative critic mechanism for diagnosing GUI operations. 
    To achieve it,
    we propose a Suggestion-aware Group Relative Policy Optimization strategy integrating a novel suggestion reward,
    and develop our pre-critic model GUI-Critic-R1.
    This model is capable of delivering constructive and insightful feedback to refine the GUI reasoning process.
    \item 
    We present a reasoning-bootstrapping based data collection pipeline to construct a GUI-Critic-Train dataset, comprising 6k high-quality chain-of-thought annotations. Additionally, a GUI-Critic-Test dataset is developed to comprehensively evaluate the critic model's performance in both mobile and web domains. 
    \item 
    Extensive experiments on both GUI-Critic-Test dataset and dynamic GUI automation benchmark validate the efficacy of our GUI-Critic-R1 model in producing reliable judgments and feedback for GUI operations.
    }
\end{enumerate}

%% file: sec/2_related_work.tex
\section{Related Work}

\label{sec:related_work}

\textbf{LLM-based GUI Agent. }
Recently, significant attention has been directed toward Graphical User Interface (GUI) agents for task automation on smart devices designed for the mobile and web environments~\cite{zhang2023appagent,li2024appagent, wang2024mobile, zhang2024ufo, zheng2025vem,yan2023gpt, hong2024cogagent, zhang2024smartagent, ma2024coco,song2024visiontasker, wen2024autodroid, wen2023droidbot}. 
{
Despite these advancements, GUI agents frequently make erroneous decisions. 
Some research has incorporated reflection modules in GUI automation frameworks to verify operation correctness based on the both current screenshot and the screenshot after execution~\cite{wang2024mobile2,liu2025pc,agashe2025agent}.
For example, Mobile-Agent-v2 frameworks~\cite{wang2024mobile} employ multi-agent architectures that separate planning, decision-making, and reflection to optimize task tracking, memory, and error correction.
But these methods require additional steps to undo actions and pose the risk of irreversible operations, resulting in lower efficiency and accuracy. 
In this paper, we introduces a pre-critic mechanism for diagnosing potential errors.
}

\textbf{Critic Model for LLMs. }
Large language models (LLMs) do not always generate the best reasoning output when performing challenging inference tasks~\cite{tyen2023llms}. 
To address this limitation, several studies propose self-refinement\cite{madaan2023self, sun2023adaplanner, pan2024autonomous} and self-reflect techniques~\cite{shinn2023reflexion, gupta2024metareflection, wang2024devil, yuan2025agent, zhang2024agent, dou2024re, fu2025agentrefine} to reifine the output themselves. 
However, their effectiveness is largely restricted by their reliance on the inherent capabilities of LLMs, which may hinder the broader application and scalability of these methods~\cite{huang2023large}. 
In contrast, several studies~\cite{mcaleese2024llm, xiong2024llava, xiang2024retrospex} explore employing an independent critic model to produce natural language feedback for the evaluation of outputs generated by Large Language Models (LLMs). 
Critic-V~\cite{zhang2024critic} extends this inspiration to the area of VLMs to train a critic vision-language model to locate imperfections in visual content perception and errors in reasoning steps. 
However, they focus on general offline tasks. 
{
Differently, we explore the application of a critic model in more complex scenarios of GUI automation, which presents increased challenges due to its operation in an online environment. 
In this paper, we perform pre-operative critic to resolve the challenges. 
}

\textbf{Reinforcement Learning for Reasoning. }
Recent research has increasingly focused on enhancing the reasoning capabilities of LLMs through Reinforcement Learning (RL), drawing inspiration from models such as DeepSeek-R1~\cite{guo2025deepseek} and Kimi-1.5~\cite{team2025kimi}. 
These models employ rule-based reward mechanisms to enhance reasoning performance.
Several studies attempt to adapt the idea of reinforcement learning with verifiable rewards in multimodal scenarios. 
For instance, R1-V~\cite{zhang2025r1} investigates the application of rule-based RL in geometry problems and object-counting tasks, while Visual-RFT~\cite{liu2025visual} extends this approach to open vocabulary and few-shot detection, reasoning grounding, and fine-grained few-shot classification. 
Recent studies~\cite{huang2025vision, peng2025lmm, meng2025mm, zhang2025r1, yang2025r1} further generalize the algorithm to address more general tasks such as multimodal mathematical reasoning, decision-making, and planning. 
In this paper, we extend the reinforcement learning algorithm to train a critic model that diagnoses operations in GUI automation, and propose a S-GRPO strategy with a novel suggestion reward to enhance the model's reasoning capabilities.

%% file: sec/4_method.tex
\section{Method}

\label{sec:method}

\begin{figure*}[t]
\centering
\includegraphics[width=1\linewidth]{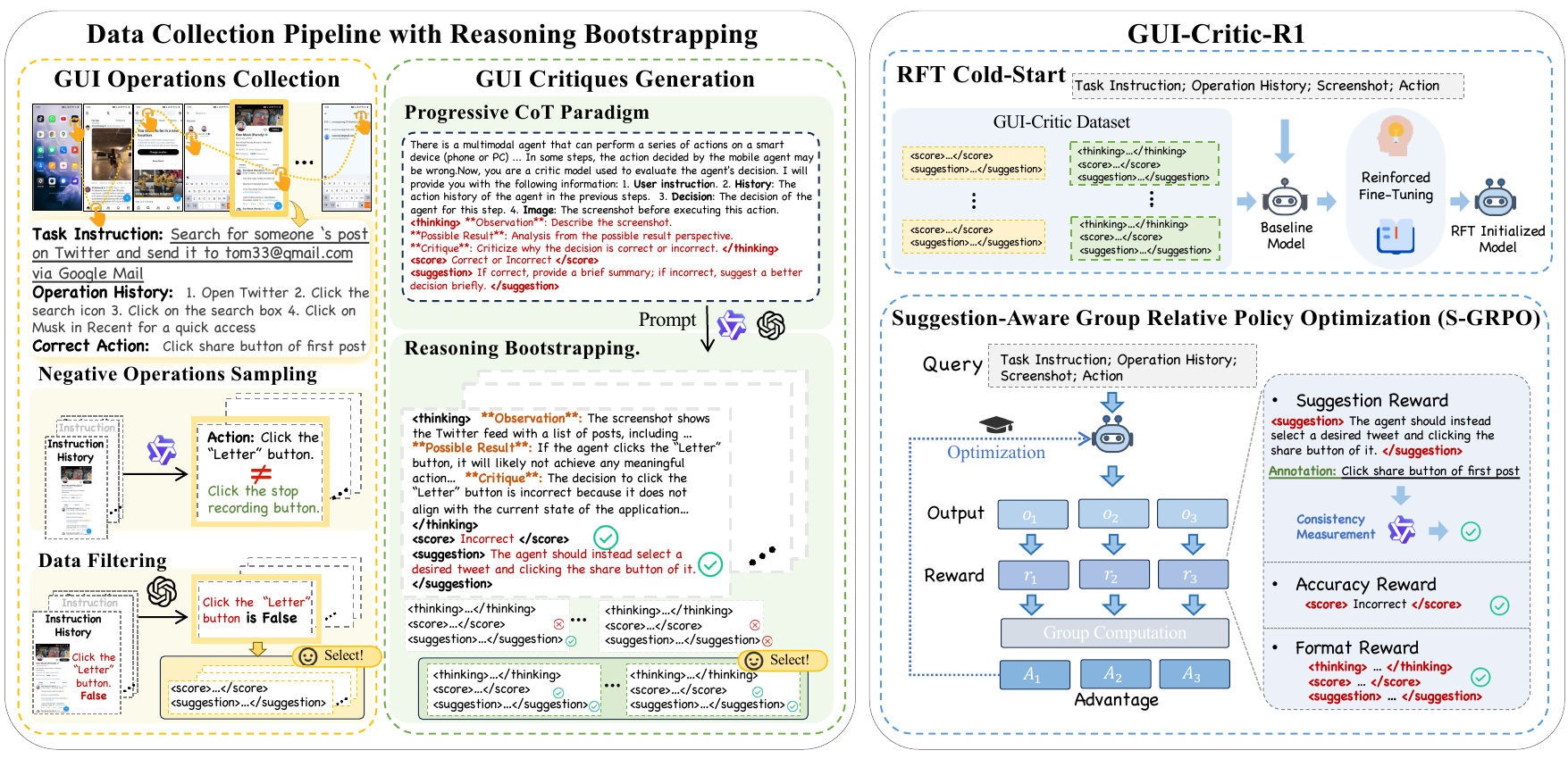}
\vspace{-4mm}
\caption{\small
The left shows our reasoning-bootstrapping based data collection pipeline,
including the GUI operations collection and GUI critiques generation.
Specifically, a progressive CoT paradigm and a reasoning-bootstrapping strategy are employed to ensure the quality of critiques.
The right illustrates the training strategy for our GUI-Critic-R1 model. 
The process begins with a RFT cold-start on the GUI-Critic-Train dataset, and followed by the implementation of our proposed S-GRPO. Besides, a novel suggestion reward is employed to constrain the correctness of suggestions.
}
\vspace{-2mm}
\label{fig2}
\end{figure*}

\subsection{Problem Definition of Pre-Operative Critic}
\label{sec: problem definition}
The GUI automation task
can be formally characterized as a Markov Decision Process: 
$\mathcal{M} = (\mathcal{E}, \mathcal{A}, \mathcal{P}, \pi_{agent})$. 
The state of the environment $\epsilon \in \mathcal{E}$ consists of a user instruction, historical interactions, and the current screenshot of the device. 
The action space $\mathcal{A}$ encompasses all available operations (actions)
including {\textit{click, long press, type, scroll, home, back}, and \textit{done}}. 
%
%
At each step, the MLLM-based agent $\pi_{agent}$ observes the environment $\epsilon$ and selects an action $a$. 
Upon executing $a$, it receives a novel observation $\epsilon'$, with the probability given by the state transition function $\mathcal{P}(\epsilon'|\epsilon, a)$.
The agent iterates this process until accomplishing the desired instruction or encountering a terminal state.

In this paper, we propose a pre-operative critic model $\pi_{critic}(\epsilon, a)$ to critique the decision-making process of $\pi_{agent}$, before the action is actually executed.
Specifically,
taking state $\epsilon$ and action $a$ as inputs, the model produces a
correctness score ${l} \in [0,1]$ that reflects the rightness of $a$.
Besides, it generates critique ${c}$ and corrective suggestion ${s}$ in natural language,
with the former explaining the rationale behind the correctness judgment and the latter suggesting a better action if $a$ is incorrect.

\vspace{-2mm}
\subsection{{Data Collection Pipeline with Reasoning Bootstrapping}}
\label{sec:data collection}
\vspace{-2mm}

In this section, we detail the data collection pipeline of our proposed GUI-Critic-Train and GUI-Critic-Test datasets. 
Each sample in the dataset includes the environmental state $\epsilon$ of the step, a operation candidate $a$, as well as annotations for correctness score $l$ and suggestion ${s}$.
This section commences by introducing the collection of operations in Section~\ref{sec:data collection:stage1}, followed by the presentation of our reasoning bootstrapping method for generating critique $c$ in Section~\ref{sec:data collection:stage2}.

\vspace{-2mm}
\subsubsection{GUI Operations Collection}
\label{sec:data collection:stage1}
To begin with, we collect successful automation trajectories from publicly accessible datasets, encompassing correct step-level operations across various GUI scenarios. 
The action in a specific state $\epsilon$ is considered as the operation candidate whose correctness score is positive ($l=1$), and the corresponding action description is considered as the suggestion ${s}$.

\textbf{Negative Operations Sampling. } 
Subsequently, we collect samples with negative score ($l=0$) based on the states of the above correct operations.
To be specific,
given these states,
open-source MLLM-based agents are first employed to predict a set of operations.
Then, these operations are evaluated according to rule-based criteria, and we retain those deemed incorrect.
In this way, 
the obtained negative operations are aligned as closely as possible with the real error distribution in GUI environments, thus ensuring the quality of the dataset.
The corrective suggestions for these samples are the same as the corresponding positive samples. 

\textbf{Data Filtering. } 
The data collected using the aforementioned methods may not be entirely reliable. This is because that public datasets could contain erroneous annotations, 
and the rule-based criteria are not entirely reliable, as they may also erroneously penalize some correct operations.
Consequently, further data cleaning is necessary.
Specifically, we adopt GPT-4o~\cite{gpt4o} as a pre-critic model to judge the rightness of the operations in the collected samples,
and we retain the samples for which the annotated scores are consistent with the judgment by GPT-4o. 
Finally, we denote the collected samples as  $\mathcal{D}_{c\_action}=\{({\epsilon}_n,a_n,l_n,s_n)\}_{n=1}^N$.

\subsubsection{GUI Critiques Generation}

\label{sec:data collection:stage2}

The $\mathcal{D}_{c\_action}$ collected in the above section still lacks the 
critique $c$ that elucidates why an action is considered as correct or incorrect. 
In this section, we resort to Chain-of-Thought (CoT) technique to enabling MLLMs to handle the challenging critique generation.

\textbf{Progressive CoT Paradigm.}
We design a progressive CoT paradigm to helps MLLMs to perform deliberate, structured, and analytical thinking about the GUI operations.
Specifically, our paradigm contains three parts \texttt{<thinking>...</thinking>
<score>...</score>
<suggestion>...</suggestion>}, which respectively corresponding to intermediate logical thought $t$, correctness score $l$, and suggestion $s$. 
The reasoning $t$ contains the content of critique $c$. 
In particular, the logical reasoning in the <thinking> field is composed of several essential components:

{\small
\begin{enumerate}[nosep] 
    \item \texttt{Observation: Analyze the screenshot's state, ignoring user instructions.}
    \item \texttt{Possible Result: Speculate the most possible result of the action.}
    \item \texttt{Critique: Assess correctness of the action with reasoning.}
\end{enumerate}
}
\texttt{Observation} enhances the model's perceptual acuity by demanding a comprehensive understanding of spatial and contextual elements present in GUI. 
\texttt{Possible Result} boosts foresight by prompting the model to predict outcomes from current observations. 
Finally, \texttt{Critique} involves a critical analysis of actions to verify their correctness. 

\textbf{Reasoning Bootstrapping.}
Following the above format, we leverage current MLLMs to generate CoT reasoning progress.
However, 
performing reverse annotation of CoT conditioned on ground-truth score $l$ and suggestion $s$ is harmful.
This is because that the MLLMs may overly depend on pre-known annotations,
potentially biasing the thought process toward these outcomes rather than reflecting the actual critique reasoning sequence. 
Therefore,
we propose a reasoning bootstrapping strategy to generate high-quality thoughts without prior knowledge of the ground-truth $l$ and $s$.
Specifically,
merely provided with environmental states $\epsilon$ and action candidates $a$,
it forces the model to simulate a genuine critic reasoning process for ${max}$ times as follows: 
\begin{equation}
  \mathbb{E}_{(\epsilon, a, l, s) \sim \mathcal{D}_{c\_action}} (\bar{t}_{i}, \bar{l}_{i}, \bar{s}_{i})^\text{max}_{i=1} = \bar{\pi}(\epsilon,a) ,
  \quad  \text{Select} \, (\bar{t}_{i},\bar{l}_{i}, \bar{s}_{i}) \mid (\bar{l}_{i} = l) \land (\bar{s}_{i} = s),
\label{equ:Bootstrapped_Thought}
\end{equation}

where $\bar{\pi}$ is an excellent MLLM and $\bar{t}$ denotes the generated thought.
We select reasoning outputs containing the correctness score and suggestion that match with annotations to compile $\mathcal{D}_{c\_cot}=\{(\epsilon_m,a_m,l_m,s_m, t_m)\}_{m=1}^M$. 

Finally, the GUI-Critic-Train dataset is constructed by aggregating the collected datasets, which can be represented as  $\mathcal{D}_{c}=\mathcal{D}_{c\_action} \cup \mathcal{D}_{c\_cot}$.
For the GUI-Critic-Test dataset that does not require the annotation of reasoning process,
we construct it with the pipeline introduced in Section~\ref{sec:data collection:stage1}.

\subsection{GUI-Critic-R1}

Constructing a pre-critic model for GUI automation is not easy since it requires a thorough understanding of GUI knowledge, multimodal processing, and logical reasoning abilities. 
In this section, to cultivate the deliberative analysis ability of the pre-critic model for complex GUI scenarios,
we propose a Suggestion-aware Group Relative Policy Optimization (S-GRPO) strategy for our GUI-Critic-R1.
Specifically,
we first employ Reinforcement Fine-Tuning Initialization (RFT cold-start) to stabilize the model's reasoning process.
Then, the S-GRPO is adopted to further enhance the model's ability for GUI pre-critic.

\subsubsection{RFT Cold-Start}

We initiate model training with Reinforced Fine-Tuning (RFT) on collected GUI-Critic dataset $\mathcal{D}_{c}$:
\begin{equation}
\small
    \mathcal{L}_{rft} = \mathbb{E}_{(\epsilon,a,l,s) \sim \mathcal{D}_{c\_action},\  (\hat{\epsilon},\hat{a},\hat{l},\hat{s},\hat{t}) \sim \mathcal{D}_{c\_cot}} -\text{log}(\pi_{critic}(l, s|\epsilon,a))-\text{log}(\pi_{critic}(\hat{l},\hat{s},\hat{t}|\hat{\epsilon},\hat{a})). 
\label{eq1}
\end{equation}
The cold-start stage distills GUI knowledge from human annotations (\textit{i.e.}, correct operations in $\mathcal{D}_{c\_action}$)
and distill reasoning experience from existing MLMs (\textit{i.e.}, progressive reasoning processes in $\mathcal{D}_{c\_cot}$),
thereby equipping the model with foundational capabilities needed for generating operation critiques and valid feedbacks.

\subsubsection{Suggestion-aware Group Relative Policy Optimization}
After the RFT cold-start, we enable the pre-critic model to self-improve its reasoning ability via online reinforcement learning.
Standard GRPO~\cite{guo2025deepseek} approach samples a group of generated outputs for each input from the policy model, where each output includes CoT thought and an answer. 
Subsequently, the model is encouraged to favor better answers with a high reward value within the group. 
Since the pre-critic model is required to output additional valuable critiques and suggestions of the operation,
only employing the typical reward that focus on the correctness of format and answer is not enough.
Therefore,
we introduce a suggestion-aware GRPO method, incorporating a novel suggestion reward specifically designed for GUI pre-critic.

\textbf{Suggestion Reward.}
Given the suggestion annotation $s'$,
the suggestion reward $r_{s}$ evaluates the model's constructive outputs as follows: 
\begin{equation}
r_{s}({o}) =\mathbb{I}_{similar}({s}, s'), 
\label{eq3}
\end{equation}
where ${o}=({l},{s},{c})$ is the output of pre-critic model $\pi_{critic}$.
$\mathbb{I}_{similar}(\cdot)$ calculates the similarity between the inputs leveraging large language models, and output a binary judgment. 
Based on Eq.~(\ref{eq3}),
$r_{s}$ provides a quantitative and objective metric to assess the correctness of suggestions,
thereby facilitating more precise evaluations of $\pi_{critic}$ efficacy.
Moreover,
this direct feedback enables models to learn from their outputs iteratively, refining their progressive CoT reasoning process to produce more accurate action critique and corrective suggestions.

\textbf{Format and Accuracy Reward.}
Additionally, we adopt the rewards introduced in DeepSeek-R1~\cite{guo2025deepseek}: format reward $r_f(o)$ and accuracy reward $r_a(o)$. 
The former evaluates whether the outputs follow the structure described in Section~\ref{sec:data collection:stage2}.
The latter evaluates whether the generated correctness score align with ground-truth. 
These rewards imposes constraints solely on the format and final answer, leaving the thought process unregulated, resulting in stimulating the model’s intrinsic ability to reason.

\textbf{Optmization.}
The final reward function $r(o)$ can be defined as: $r(o) = \lambda_f \cdot r_f(o) + \lambda_{s} \cdot r_{s}(o) + (1-\lambda_{f}-\lambda_{s}) \cdot r_a(o)$, where $\lambda_f$ and $\lambda_{s}$ adjust the importance of the format and suggestion rewards. 
For each input $(\epsilon, a)$, we first sample a group of generated outputs $G=\{o_1,o_2,\cdots,o_G\}$ from the frozen policy model $\pi_{\theta_{old}}$, where each output $o=(c,l,s)$ includes {critique $c$}, score $l$ and the suggestion $s$. 
Then we maximize the following objective and optimizes the critic model $\pi_{critic}$: 

\vspace{-2mm}
\begin{equation}
\small
\mathcal{L}_{\mathrm{GRPO}}
= 
  \frac{1}{|G|} \sum_{i=1}^G
  \min\!\Bigl(
    \frac{\pi_{critic}(o_i \mid \epsilon,a)}{\pi_{\theta_{\mathrm{old}}}(o_i \mid \epsilon,a)}\,A_i,\, \\
    \mathrm{clip}\!\Bigl(
      \frac{\pi_{critic}(o_i \mid \epsilon,a)}{\pi_{\theta_{\mathrm{old}}}(o_i \mid \epsilon,a)},
      1-\varepsilon,\,
      1+\varepsilon
    \Bigr)
    A_i
  \Bigr)
  -\,\beta\,D\bigl(\pi_{critic}\,\big\|\,\pi_{\mathrm{ref}}\bigr),
\label{eq1}
\end{equation}
where $\varepsilon$ and $\beta$ are the PPO clipping hyperparameters and the coefficient controlling the Kullback–Leibler (KL) penalty, respectively. 
$D\bigl(\pi_{critic} \,\|\, \pi_{\text{ref}}\bigr)
= \frac{\pi_{\text{ref}}(o_i \mid \epsilon,a)}{\pi_{critic}(o_i \mid \epsilon,a)}
- \log\!\Bigl(\frac{\pi_{\text{ref}}(o_i \mid \epsilon,a)}{\pi_{critic}(o_i \mid \epsilon,a)}\Bigr)
- 1 \,$ is the KL divergence. 
The relative advantage for the $i$-th response is computed by normalizing the rewards across the group:
$A_{i} = \frac{r(o_i) - \text{mean}(\{r(o_1), \dots, r(o_G)\})}{\text{std}(\{r(o_1), \dots, r(o_G)\})}$. 

%% file: sec/5_experiment.tex
\section{Experiment}
\label{sec:experiment}

\subsection{Experiment Settings}
\label{sec:experiment:setting}
\textbf{Dataset and Benchmark.}
\textbf{\textit{(1) {GUI-Critic-Train.}}}
The data in GUI-Critic-Train were sourced from publicly available GUI operation datasets, including AITZ~\cite{zhang2024android}, AMEX~\cite{chai2024amex}, Odyssey~\cite{lu2024gui}, and AITW~\cite{rawles2023androidinthewild}. 
We utilize Qwen2.5-VL-7B~\cite{bai2025qwen2} to generate negative operations, which were simply deemed incorrect based on rule-based criteria. 
GPT-4o~\cite{gpt4o} and Qwen2.5-VL-72B~\cite{bai2025qwen2} are utilized to generate the Chain of Thought (CoT) processes for {\color{black}$\mathcal{D}_{c\_cot}$} outlined in Section~\ref{sec:data collection:stage2}. 
{\color{black}
Consequently, GUI-Critic-Train dataset comprises approximately 11k entries, with 6k annotated by high-quality Chain-of-Thought (CoT) processes. }
\textbf{\textit{(2) {GUI-Critic-Test.}}}
To comprehensively evaluate the pre-critic capabilities of MLLMs, 
we established three main benchmark settings in GUI-Critic-Test: Mobile-Instruction Generalization (\textbf{GUI-I}), Mobile-Scenario Generalization (\textbf{GUI-S}), and Web-Scenario Generalization (\textbf{GUI-W}).
Specifically,
GUI-I test data are sourced from the AMEX~\cite{chai2024amex}, with instructions different from those in the training set. 
GUI-S test data are drawn from the Odyssey~\cite{lu2024gui}, comprising mobile applications that differ from those seen in the training set. 
Besides, shifting from mobile to web, GUI-Web contains the samples about the web automation, which are randomly sampled from the GUICourse~\cite{chen2024guicourse}. 
Each of the settings, GUI-I, GUI-S, and GUI-Web, is manually annotated to ensure label accuracy, resulting in dataset sizes of 656, 114, and 418, respectively. 
We present critic accuracy and suggestion accuracy as the metrics.
{\color{black}
The former reflects the ability to assess the correctness of GUI operations, while the latter reflects the suggestion quality by quantifying the similarity between the generated suggestion and the annotation, which is calculated by prompting the Qwen2.5-VL-72B \cite{bai2025qwen2}. 
}

\textbf{Implementation Details.}
We employ Qwen2.5-VL-7B~\cite{bai2025qwen2} as the backbone. 
{\color{black}
For the RFT cold-start, it is conducted for one epoch on mobile scenarios (GUI-I and GUI-S) and two epochs on web scenarios (GUI-W) via supervised fine-tuning (SFT). 
}
Next, we train the model on the $\mathcal{D}_{c\_action}$ by S-GRPO for 10 epochs, with a learning rate of $3 \text{e}^{-6}$ and a batch size of 128. 
The suggestion and format reward weights ($\lambda_{s}$, $\lambda_f$) are both set to 0.1, while the group size is set to 6. 
In accordance with the methodology outlined in~\cite{zheng2025easyr1}, the KL divergence coefficient is set to $1\text{e}^{-2}$ by default. 
All experiments are conducted on eight NVIDIA A100 Tensor Core GPUs.

\begin{table}[t]
\footnotesize
\centering
\caption{
Static evaluation performance comparison of closed-source and open-source MLLMs on our GUI-Critic-Test, including three different settings: GUI-I, GUI-S, and GUI-W. 
}
\label{tab:critic result}
\begin{tabular}{lcccccc} 
    \toprule
  \textbf{Model}& \multicolumn{2}{c}{\textbf{GUI-I}}  & \multicolumn{2}{c}{\textbf{GUI-S}} & \multicolumn{2}{c}{\textbf{GUI-W}} \\ 
  \cmidrule(l){2-3}\cmidrule(l){4-5}\cmidrule(l){6-7}& \makecell[c]{Critic\\Accuracy} & \makecell[c]{Suggestion\\Accuracy}  &\makecell[c]{Critic\\Accuracy} & \makecell[c]{Suggestion\\Accuracy} & \makecell[c]{Critic\\Accuracy} & \makecell[c]{Suggestion\\Accuracy}  \\ \midrule
\multicolumn{7}{c}{\textbf{Close Source MLMs}}  \\ \midrule
Claude-3.5  & {67.26} & 40.71 & 64.27 & 46.11 & 65.55 & 37.64 \\
GPT-4o  & 66.01  & 40.54  & 62.28  & 33.33 & 68.45  & 28.27 \\
Gemini-2.0-Flash  & 66.76 & 42.98 & 64.91 & 38.59 & 62.85 & 38.78  \\ \midrule
\multicolumn{7}{c}{\textbf{Open Source MLMs}}  \\\midrule
Deepseek-VL2-7B~\cite{wu2024deepseekvl} & 44.36  & 0.00  & 43.85  & 0.00 & 10.28  & 0.00 \\
InternVL2.5-8B~\cite{chen2024expanding}  & 53.96  & 19.35  & 52.63  & 14.91 & 54.67  & 24.53 \\
InternVL2.5-8B-MPO~\cite{chen2024expanding}  & 51.06  & 20.42  & 51.75  & 19.30 & 55.37  & 24.06 \\
Qwen2.0-VL-7B~\cite{wang2024qwen2}  & 52.59  &  21.04  & 54.42  &  19.30 & 53.50 & 38.78\\
Qwen2.0-VL-72B~\cite{wang2024qwen2}  & 54.27  & 30.03  & 53.51  & 29.80 & 51.86  & 21.96 \\
Qwen2.5-VL-7B~\cite{bai2025qwen2}  & 54.88  & 43.14  & 57.02  & 37.72 & 59.11  & 36.21 \\
Qwen2.5-VL-72B~\cite{bai2025qwen2}  & \underline{56.40}  & \underline{49.08}  & \textbf{59.65} & \underline{38.79} & \underline{60.05}  & \underline{38.79} \\
\midrule
\textbf{Ours  (GUI-Critic-R1)}  & \textbf{69.20} & \textbf{52.43} & \underline{58.77} & \textbf{47.37} & \textbf{63.08} & \textbf{39.48}\\ 
\rowcolor{cyan!10} $\Delta$~(Ours - Qwen2.5-VL-7B) & +14.32 & +9.29 & +1.75 & +9.65 & +3.97 & +3.27\\
\bottomrule
\end{tabular}
\label{tab:evaluation}
\vspace{-2mm}
\end{table}

\subsection{Comparision Results}
To analyze the performance of our GUI-Critic-R1 comprehensively, we construct experiments from both static and dynamic aspects. 
Firstly, we evaluate the model's ability to determine operational correctness and suggest the corrective operation on our static GUI-Critic-Test benchmark. 
Additionally, a crucial application of GUI-Critic lies in its functionality as a pre-critic model within the dynamic GUI automation processes.
To this end, we implement online evaluation experiments on AndroidWorld~\cite{rawles2024androidworld} to validate the effectiveness of our approach. 

\subsubsection{Static Evaluation}
Table~\ref{tab:evaluation} illustrates the quantitative results of our GUI-Critic-R1 and a variety of baselines on GUI-Critic-Test benchmark.
It showcases that GUI-Critic-R1 performs competitively on different scenarios compared to closed- and open-source MLMs in both critic accuracy and suggestion accuracy. 
For close source MLMs, {\color{black}Claude-3.5} achieves best performance on most settings.
Despite the capabilities of close source MLMs, our GUI-Critic-R1 achieves state-of-the-art (SoTA) performance on the GUI-I test dataset. 
Compared to the base model, GUI-Critic-R1 achieved a significant improvement in critic accuracy from 54.88\% to 69.05\%. 
Compared to the excellent GPT-4o, our model also achieves a 3.19\% critic accuracy improvement and a 11.89\% advantage in suggestion accuracy. 
When faced with GUI-S, the novel application scenarios containing complex cross-application instructions from Odyssey~\cite{lu2024gui}, our model demonstrated commendable generalization capabilities as well. 
Although our model achieves a relatively insignificant improvement in critic accuracy (1.75\%), it performs a 9.65\% advantage in suggestion accuracy. 
It demonstrates that our model has a robust understanding of operations even in a new environment. 
In the web scenarios, where domain differences are more significant, our model exhibits commendable performance, indicating that it not only enhances the ability to determine the operation correctness but also generates effective suggestions. 
These results validate the superiority and robustness of our proposed S-GRPO.

\begin{wraptable}{r}{0.4\textwidth}
    \vspace{-10mm}
      \caption{ 
      Dynamic evaluation results on the AndroidWorld~\cite{rawles2024androidworld} benchmark.
    }
      \centering
      \label{tab: android world}
      \resizebox{0.99\linewidth}{!}{
      \begin{tabular}{lcc}
            \toprule
           \textbf{Model} & SR($\uparrow$) & EAR($\uparrow$)  \\
           \midrule Baseline & 22.4 & - \\
           \hline \multicolumn{3}{c}{\it{Post-Critic}} \\
           \hline GPT-4o~\cite{gpt4o} & 25.0 & 18.3 \\
           \hline \multicolumn{3}{c}{\it{Pre-Critic}} \\
           \hline
           Qwen2.5-vl-7B~\cite{bai2025qwen2} & 20.3 & 21.8 \\
           Qwen2.5-vl-72B~\cite{bai2025qwen2}& 23.2 & 24.4 \\
            GPT-4o~\cite{gpt4o} & 22.4 & 26.1 \\
           \midrule \textbf{Ours} & 27.6 & 31.8 \\
           \bottomrule
      \end{tabular}
    }
    \vspace{-2mm}
\end{wraptable}

\subsubsection{Dynamic Evaluation}
\vspace{-1mm}
We further evaluate our model on the AndroidWorld~\cite{rawles2024androidworld} benchmark, which provides a live Android emulator and 116 tasks across 20 mobile apps. 
Specifically, in this platform, the GUI agent can perform operations on an emulated Android phone to attain human instructions, and the results are evaluated automatically. 
Our pre-critic model can serve as an error diagnosis module in the framework in a plug-and-play manner. 
For a fair comparison, we conduct both post-operative and pre-operative critic with existing MLLMs as the baseline. 
{\color{black}
The results in Table~\ref{tab: android world} illustrate that GUI-Critic-R1 achieves the best success rate on the benchmark, verifying the ability of error correction and suggestion of our model. 
}
We observe that both post- and pre-critic can improve the performance of the GUI agent,  indicating that pre-critic can avoid some potential mistakes, and post-critic can also remedy some accomplished errors. 
Despite the capabilities of GPT-4o, we find that it can sometimes produce inaccurate critiques and suggestions to mislead the agent when performing as the pre-critic, primarily due to its insufficient common-sense knowledge of the GUI.
{\color{black}
GPT-4o outperforms the prior-critic on the post-critic because the prior-critic requires the model to predict the potential outcomes of GUI operations, which GPT-4o lacks the common-sense reasoning ability. }
Furthermore, we evaluated the efficiency of GUI task execution across different approaches. 
We introduced a metric called {\color{black}Efficiency Advantage Rate (EAR)} for fair comparison, which measures the proportion that the `baseline + critic' model achieves an efficiency advantage (fewer steps) than the baseline, for tasks where they achieve consistent results. 
The results shown in Table~\ref{tab: android world} reveal that our model tends to complete tasks in fewer steps, thereby demonstrating its efficiency.
In contrast, post-critic methods typically require a greater number of steps.

\begin{table}[!t]
    \begin{minipage}[t]{0.45\linewidth}
      \caption{Ablation study on the dataset collection pipeline. }
\setlength{\extrarowheight}{0pt}
\setlength{\tabcolsep}{1pt}
      \label{tab:ablation_dataset}
      \vspace{-1mm}
      \centering
      \label{ablation_data_methods}
    \resizebox{0.98\linewidth}{!}{
      \begin{tabular}{lcccc}
    \toprule
       \multirow{2}[0]{*}{Model} & \multicolumn{2}{c}{GUI-I} & \multicolumn{2}{c}{GUI-S}  \\ 
      \cmidrule(l){2-3}\cmidrule(l){4-5}& \makecell[c]{Critic\\Accuracy} & \makecell[c]{Suggestion\\Accuracy} & \makecell[c]{Critic\\Accuracy} & \makecell[c]{Suggestion\\Accuracy}  \\         \midrule
        \midrule
      w/o NOS & 50.46 & 1.22  & 50.30 & 5.26  \\
      w/o DF & 67.23 & 49.54 & 54.39 & 42.98 \\
      w/o GCG & 67.84 & 51.22 & 56.14 & 42.11 \\ 
      \textbf{Ours}  & \textbf{69.20} & \textbf{52.43} & \textbf{58.77} & \textbf{47.37} \\
    \bottomrule
      \end{tabular}}
    \end{minipage}
    \begin{minipage}[t]{0.55\linewidth}
\setlength{\extrarowheight}{0pt}
\setlength{\tabcolsep}{3pt}
      \caption{Ablation study on training strategies.}
      \centering
      \label{tab:ablation_train}
    \resizebox{0.98\linewidth}{!}{
      \begin{tabular}{cccccccc}
    \toprule
      \multirow{2}[0]{*}{RFT}& \multicolumn{3}{c}{S-GRPO}& \multicolumn{2}{c}{GUI-I} & \multicolumn{2}{c}{GUI-S} \\         
      \cmidrule(l){2-4}\cmidrule(l){5-6}\cmidrule(l){7-8} & $r_{f}$ & $r_{a}$ & $r_{s}$ & \makecell[c]{Critic\\Accuracy} & \makecell[c]{Suggestion\\Accuracy} & \makecell[c]{Critic\\Accuracy} & \makecell[c]{Suggestion\\Accuracy}  \\ \midrule
        \midrule
      \ding{51} & \ding{55} & \ding{55} & \ding{55} & 63.16 & 45.61 & 55.26 & 34.21 \\
      \ding{55} & \ding{51} & \ding{51} & \ding{55} & 67.98 & 43.44 & 54.38 & 39.47 \\ 
      \ding{55} & \ding{51} & \ding{51} & \ding{51} & 69.05 & 49.24 & 56.14 & 42.10 \\ 
      \ding{51} & \ding{51} & \ding{51} & \ding{55} & 66.01 & 47.71 & 57.89 & 40.35 \\ 
      \ding{51} & \ding{51} & \ding{51} & \ding{51} & \textbf{69.20} & \textbf{52.43} & \textbf{58.77} & \textbf{47.37} \\
    \bottomrule
      \end{tabular}}
        \end{minipage}
\end{table}

\begin{figure}[!t]
    \begin{minipage}{0.4\linewidth}
        \centering
        \includegraphics[width=0.98\linewidth]{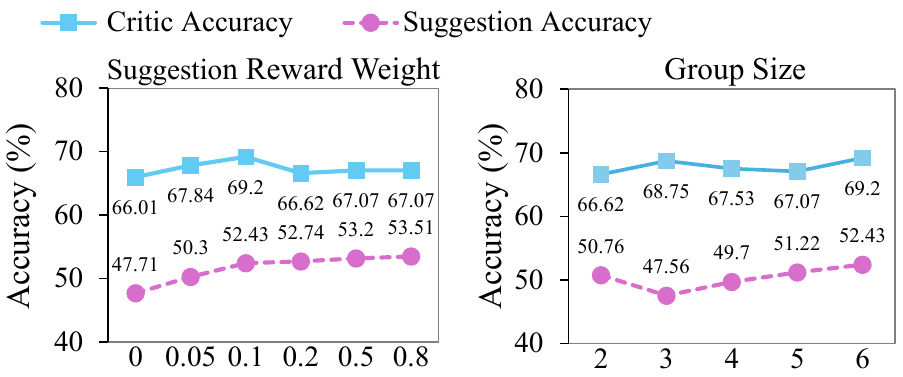}
        \caption{
        Analysis of suggestion reward weight $\lambda_{s}$ and  Group Size on the GUI-I setting.}
        \label{fig: parameters}
    \end{minipage}
        \hfill
    \begin{minipage}{0.58\linewidth}
    \centering
    \vspace{-2mm}
    \includegraphics[width=0.98\linewidth]{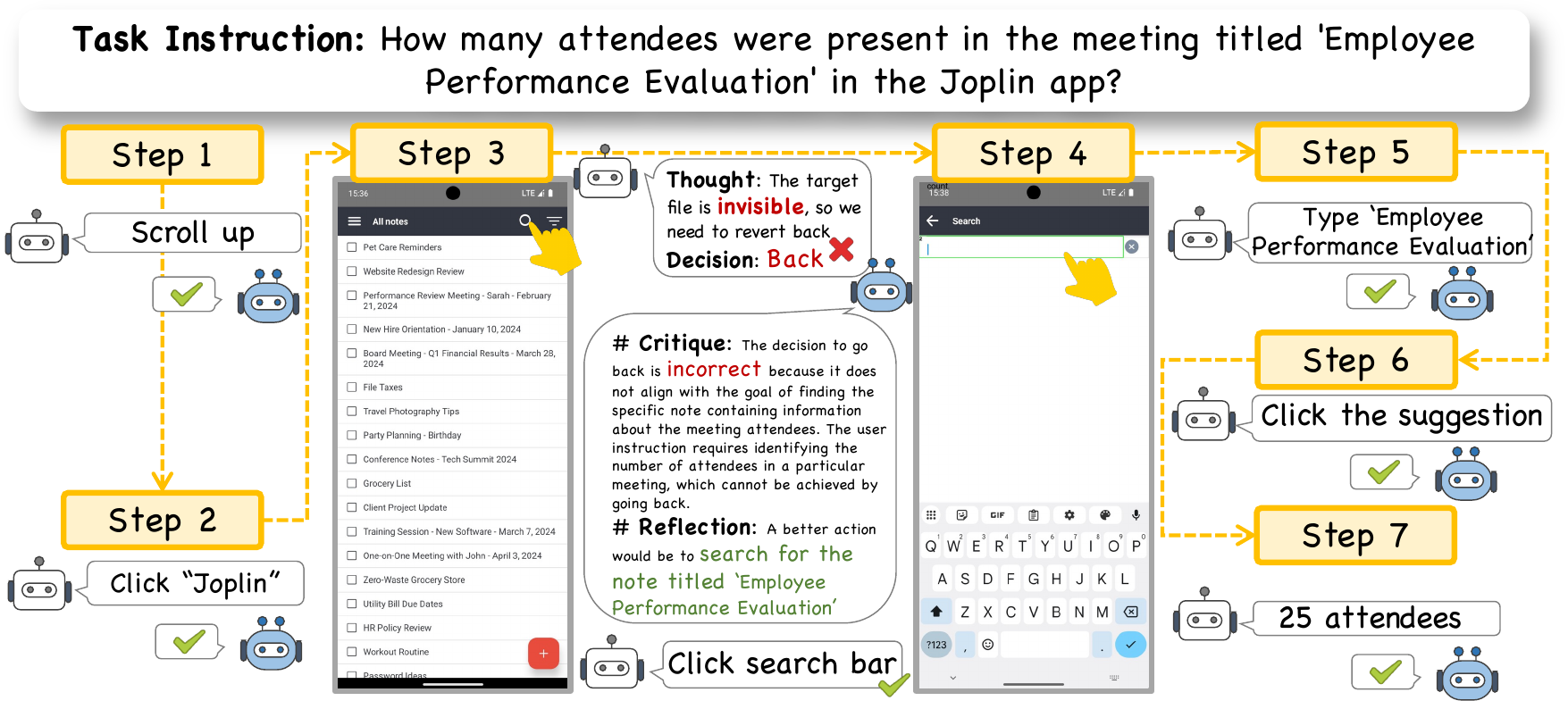}
    \vspace{-2mm}
     \caption{A case of pre-critic on GUI automation. }
         \label{fig:case}
    \end{minipage}
    \vspace{-3mm}
\end{figure}

\subsection{Ablation Study}
\vspace{-1mm}
\textbf{Analysis of Dataset Collection Pipeline. }
We first conduct ablation studies for {\color{black} the proposed data collection pipeline} to verify the contributions of three key components on GUI-I and GUI-S settings. 
{\color{black} Corresponding to  Table~\ref{tab:ablation_dataset}}, we first substitute Negative Operations Sampling (NOS) with a strategy of random decision replacements to acquire negative samples. 
It underscores that negative samples sampled by random substitution are too naive to undermine the model's capability to detect errors and suggest corrections. 
Secondly, we substitute the data filtering (DF) phase with a random sample selection approach. 
Without DF, the quality of training samples may be compromised, potentially degrading the effectiveness of the training process. 
Finally, we omit the GUI Critique Generation (GCG) process by utilizing only $\mathcal{D}_{c\_action}$ in the RFT cold-start stage. 
The absence of GCG leads to a 1.36\% decline in critic accuracy on GUI-I, highlighting its crucial contribution to the model's generalization and cognitive capabilities. 
These findings collectively emphasize the pivotal importance of our data collection process, establishing it as an integrated mechanism for boosting model training and operational efficiency.

\textbf{Analysis of Training Strategy.}
In order to analyze the impact of training strategies on model performance, we conduct ablation experiments focusing on the RFT cold-start and reward components in the GRPO stage. 
As depicted in Table~\ref{tab:ablation_train}, the results in the first and third lines indicate that the utilization of RFT cold-start provided a robust initial boost to the model's decision-making capabilities, and S-GRPO further increases the ability. 
{\color{black}
Towards the second line, we observe that employing GRPO alone was insufficient to activate the model's GUI-Critic abilities, as the model does not yet possess the basic GUI critic ability, making it difficult to produce high-quality outputs when sampling randomly. 
}
In the fourth line, we ablate the suggestion reward, which led to an obvious decrease in suggestion accuracy, indicating this reward is necessary for the model to suggest correct GUI operations. 
These observations underscore the crucial significance of the RFT cold-start and the proposed suggestion reward in S-GRPO, incrementally enhancing the capability of our GUI-Critic-R1. 

\textbf{Analysis of parameters.}
Figure~\ref{fig: parameters} illustrates the impact of different suggestion reward weight $\lambda_{s}$ or the group size in the R-GRPO phase. 
Note that when varying $\lambda_{s}$, the weights for $r_{a}$ and $r_{f}$ are maintained at a ratio of 4:1, ensuring that the total sum of weights remains equal to 1. 
By adjusting $\lambda_{s}$, we observe that excessively low values inadequately constrain the suggestion, while excessively high values may compromise accuracy. 
Consequently, a value of 0.1 was selected as an optimal parameter. 
Additionally, the size of the group plays a crucial role in balancing performance with resource utilization. 
Therefore, a compromise was made by selecting a group size of 6.


\begin{figure*}[t]
    \centering
    \includegraphics[width=0.97\linewidth]{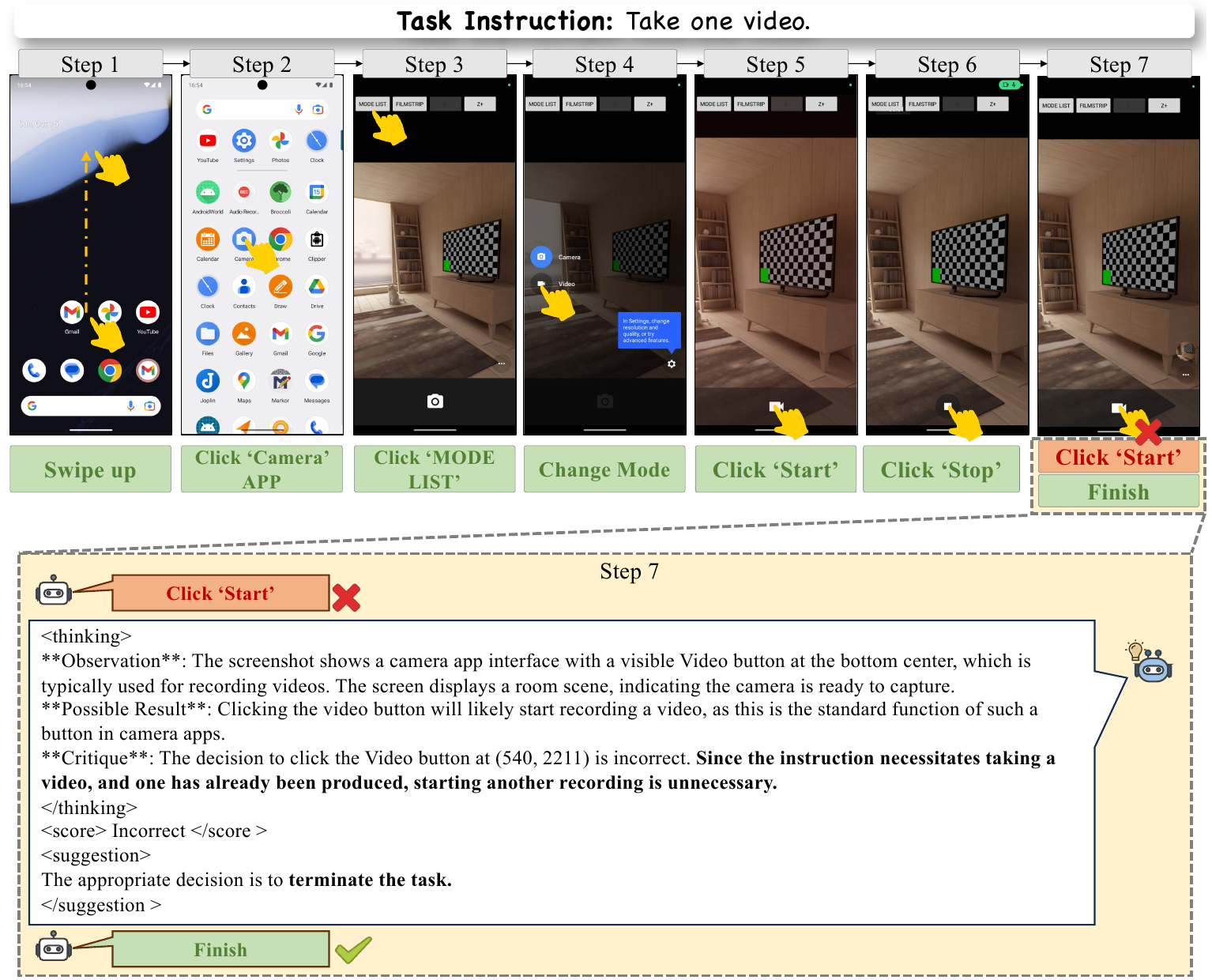}
    \vspace{-2mm}
    \caption{An instance of GUI-Critic-R1 avoiding a redundant action within GUI automation through a pre-operative critique approach.}
    \vspace{-5mm}
    \label{fig:more_case2}
\end{figure*}

\subsection{Case Study}

Figure~\ref{fig:case} illustrates an example drawn from the AndroidWorld benchmark~\cite{rawles2024androidworld}, demonstrating how our model guides the GUI agent to achieve a correct and effective trajectory. 
Specifically, the agent is tasked with finding a file in the Joplin app but encounters an interface without the target file visible, prompting it to mistakenly consider rolling back. 
Our model suggests clicking the search box to locate the target file, thereby assisting the agent in successfully completing the task. 


In Figure~\ref{fig:more_case2}, we depict another example of pre-operative critic for GUI automation. 
The agent successfully navigates steps 1 through 6, initiating the camera app and completing a video recording. 
Nonetheless, at step 7, the agent erroneously decides to press the record button again. 
This action is incorrect, as the instruction clearly stipulates recording only one video. 
Given that the agent has fulfilled this requirement, the task should be stopped rather than continuing the recording. 
Our model precisely delineates the state of the current interface and, based on historical operation, identifies the agent's decision as redundant. 
Consequently, our model advises the agent to terminate the task at this step.

%% file: sec/6_conclusion.tex
\vspace{-2mm}
\section{Conclusion}
\vspace{-3mm}
\label{sec:conclusion}
In this paper,
we introduce a pre-operative critic mechanism aimed at performing error diagnosis in GUI Automation,
prior to executing actions within interactive environments.
To develop our pre-critic model GUI-Critic-R1, 
we propose a Suggestion-aware Gradient Policy Optimization (S-GRPO) strategy with a novel suggestion reward.
It is designed to ensure effective corrective discrimination for operations and provides reliable feedback for improvement. 
Futhermore, 
We develop a reasoning-bootstrapping based data collection pipeline to construct a GUI-Critic-Train and GUI-Critic-Test dataset, addressing existing gaps in GUI critic data.
Through extensive static and dynamic experiments, we demonstrate the
the efficacy of our approach across diverse GUI scenarios, highlighting its potential to significantly enhance the success rate of automated tasks.

%% file: sec/checklist.tex
\clearpage
\section*{NeurIPS Paper Checklist}
\begin{enumerate}
\item {\bf Claims}
    \item[] Question: Do the main claims made in the abstract and introduction accurately reflect the paper's contributions and scope?
    \item[] Answer: \answerYes{}
    \item[] Justification: The abstract and introduction clearly state the paper’s contributions and scope.

\item {\bf Limitations}
    \item[] Question: Does the paper discuss the limitations of the work performed by the authors?
    \item[] Answer: \answerYes{}
    \item[] Justification: The limitations are discussed in the Appendix.

\item {\bf Theory Assumptions and Proofs}
    \item[] Question: For each theoretical result, does the paper provide the full set of assumptions and a complete (and correct) proof?
    \item[] Answer: \answerYes{}

    \item {\bf Experimental Result Reproducibility}
    \item[] Question: Does the paper fully disclose all the information needed to reproduce the main experimental results of the paper to the extent that it affects the main claims and/or conclusions of the paper (regardless of whether the code and data are provided or not)?
    \item[] Answer: \answerYes{} 
    \item[] Justification: We have provided the necessary information for reproduction in Section~\ref{sec:experiment-setup}.

\item {\bf Open access to data and code}
    \item[] Question: Does the paper provide open access to the data and code, with sufficient instructions to faithfully reproduce the main experimental results, as described in supplemental material?
    \item[] Answer: \answerNo{}
    \item[] Justification: Justification: The codes for reproducing all experimental results will be released as soon as the paper is accepted by NeurIPS.

\item {\bf Experimental Setting/Details}
    \item[] Question: Does the paper specify all the training and test details (e.g., data splits, hyperparameters, how they were chosen, type of optimizer, etc.) necessary to understand the results?
    \item[] Answer: \answerYes{}
    \item[] Justification: We have provided the training and test details necessary to understand the results in Section~\ref{sec:experiment:setting}.

\item {\bf Experiment Statistical Significance}
    \item[] Question: Does the paper report error bars suitably and correctly defined or other appropriate information about the statistical significance of the experiments?
    \item[] Answer: \answerYes{}

\item {\bf Experiments Compute Resources}
    \item[] Question: For each experiment, does the paper provide sufficient information on the computer resources (type of compute workers, memory, time of execution) needed to reproduce the experiments?
    \item[] Answer: \answerYes{}
    \item[] Justification: We provide the compute workers details in Section~\ref{sec:experiment:setting}, and the amount of compute are illustrated in Appendix.

\item {\bf Code Of Ethics}
    \item[] Question: Does the research conducted in the paper conform, in every respect, with the NeurIPS Code of Ethics \url{https://neurips.cc/public/EthicsGuidelines}?
    \item[] Answer: \answerYes{}
    \item[] Justification: The research conforms with the NeurIPS Code of Ethics in every respect.

\item {\bf Broader Impacts}
    \item[] Question: Does the paper discuss both potential positive societal impacts and negative societal impacts of the work performed?
    \item[] Answer: \answerNA{}
    \item[] Justification: The paper prioritizes discussing technical methodologies and results rather than societal impacts.

\item {\bf Safeguards}
    \item[] Question: Does the paper describe safeguards that have been put in place for responsible release of data or models that have a high risk for misuse (e.g., pretrained language models, image generators, or scraped datasets)?
    \item[] Answer:  \answerNA{}
    \item[] Justification: The paper exploring GUI automation on public datasets poses no such risks.

\item {\bf Licenses for existing assets}
    \item[] Question: Are the creators or original owners of assets (e.g., code, data, models), used in the paper, properly credited and are the license and terms of use explicitly mentioned and properly respected?
    \item[] Answer: \answerYes{}
    \item[] Justification: We have properly credited the used assets. All the datasets introduced in this paper are available for free to researchers for non-commercial use.

\item {\bf New Assets}
    \item[] Question: Are new assets introduced in the paper well documented and is the documentation provided alongside the assets?
    \item[] Answer: \answerNA{}
    \item[] Justification: The codes will be released once upon the paper’s acceptance.

\item {\bf Crowdsourcing and Research with Human Subjects}
    \item[] Question: For crowdsourcing experiments and research with human subjects, does the paper include the full text of instructions given to participants and screenshots, if applicable, as well as details about compensation (if any)? 
    \item[] Answer: \answerNA{}
    \item[] Justification: The paper does not involve crowdsourcing nor research with human subjects.

\item {\bf Institutional Review Board (IRB) Approvals or Equivalent for Research with Human Subjects}
    \item[] Question: Does the paper describe potential risks incurred by study participants, whether such risks were disclosed to the subjects, and whether Institutional Review Board (IRB) approvals (or an equivalent approval/review based on the requirements of your country or institution) were obtained?
    \item[] Answer:  \answerNA{}
    \item[] Justification: The paper does not involve crowdsourcing nor research with human subjects.

\item {\bf Declaration of LLM usage}
    \item[] Question: Does the paper describe the usage of LLMs if it is an important, original, or non-standard component of the core methods in this research? Note that if the LLM is used only for writing, editing, or formatting purposes and does not impact the core methodology, scientific rigorousness, or originality of the research, declaration is not required.
    \item[] Answer: \answerYes{}.
    \item[] Justification: The paper uses LLM and describes in detail the type of model used (Sec~\ref{sec:experiment:setting}) and prompt in Appendix.

\end{enumerate}

%% file: sec/appendix.tex
\clearpage
\section*{Appendix}
\appendix
\renewcommand{\appendixname}{\appendixname~\Alph{section}}

\section{Limitations and Future Works}
In this paper, we explore the pre-operative critic mechanism for GUI automation to enhance success rates and decrease the number of operational steps. 
Future endeavors have the potential to extend our method to lighter models, such as Qwen2.5-VL-3B, to achieve more efficient and improved critic performance. 
Furthermore, our algorithm is based on single-step GUI visual information and semantic operation history. 
Our approach can be enhanced by integrating the trajectory-level critic in future studies, which involves taking a sequence of screenshots as input. 
It has the potential to provide more comprehensive insights.

\section{More Dynamic Evaluation}

In this paper, we conduct the dynamic evaluation on AndroidWorld~\cite{rawles2024androidworld} and illustrate the experiment results in the main text (Section 4.2.2). 
In this section, we introduce more details of the implementation of the experiments and more experimental results.

\subsection{\color{black}Supplementary Implementation Detials}

We adopt the M3A (Multimodal Autonomous Agent for Android) framework introduced in AndroidWorld~\cite{rawles2024androidworld} as the backbone. 
In the M3A framework, the decision agent is provided with a list of available action types, guidelines for operating the phone, and a list of UI elements derived from the leaf nodes of the Android accessibility tree. 
The agent receives the current screenshot along with a Set-of-Mark annotated screenshot, which includes bounding boxes with numeric labels at the top-left corner of each UI element. 
At this stage, the agent attempts to execute the generated action, referencing specific marks. 
Prior to action execution, we insert a pre-operative critic to evaluate whether the agent's proposed action is conducive to achieving the instruction. 
If the action is deemed correct, it proceeds to execution as normal. 
Conversely, if the action is considered incorrect, we input the critic model's analytical recommendations to the decision agent, prompting it to reassess and formulate a new decision. 
After action execution, another agent is adopted to deliver a concise summary following the execution of the action. 
We leverage such semantic summaries to serve as a record of the action history. 
We follow the Androidworld to configure the UI element detection procedure and the definition of the action space. 

\subsection{Supplementary Experiments}
\begin{wraptable}{r}{0.45\textwidth}
    \vspace{-8mm}
      \caption{
      Supplementary dynamic evaluation results on the AndroidWorld~\cite{rawles2024androidworld} benchmark with GPT-4o~\cite{gpt4o} as the baseline.
    }
      \centering
      \label{tab: android world gpt4o}
      \resizebox{0.99\linewidth}{!}{
      \begin{tabular}{lcc}
            \toprule
           \textbf{Model} & SR($\uparrow$) & EAR($\uparrow$)  \\
           \midrule 
           \multicolumn{3}{c}{\it{Baselines}} \\
           \midrule
           Qwen2.5-VL-72B~\cite{bai2025qwen2} & 22.4 & - \\ 
           GPT-4o~\cite{gpt4o} & 23.9 & - \\
           \midrule \multicolumn{3}{c}{\it{GPT4o+Post-Critic}} \\
           \midrule GPT-4o~\cite{gpt4o} & 26.4 & 31.6 \\
           \midrule \multicolumn{3}{c}{\it{GPT4o+Pre-Critic}} \\
           \midrule
           Qwen2.5-VL-7B~\cite{bai2025qwen2} & 16.9 & 32.4 \\
           Qwen2.5-VL-72B~\cite{bai2025qwen2}& 25.9 & 36.8 \\
            GPT-4o~\cite{gpt4o} & 24.0 & 43.1 \\
           \midrule \textbf{Ours} & \textbf{29.4} & \textbf{46.2} \\
           \bottomrule
      \end{tabular}
    }
    \vspace{-2mm}
\end{wraptable}

We extend the dynamic evaluation experiment to incorporate GPT-4o~\cite{gpt4o} as the decision agent. 
As evidenced in Table~\ref{tab: android world gpt4o}, employing GPT-4o~\cite{gpt4o} as the decision-making agent yields performance benefits over Qwen2.5-VL-72B~\cite {bai2025qwen2}, underscoring GPT-4o's robust perception and reasoning capabilities. 
The pre-operative critic improves accuracy and exemplifies the efficacy of the baseline in detecting erroneous decisions and offering constructive improvement suggestions. 
In comparison with other pre-operative models, our GUI-Critic-R1 exhibits superior performance outcomes. 
When utilizing GPT-4o as the baseline, the advantage of the pre-critic in terms of the Efficient Advantage Rate (EAR) becomes more pronounced. 
GPT-4o tends to engage in excessive exploration during GUI automation, often opting to attempt further actions rather than terminating early, thereby resulting in a greater number of steps. 
In contrast, the pre-critic effectively anticipates erroneous or redundant exploration, thereby reducing unnecessary steps and manifesting a superior EAR metric.

\begin{figure*}[t]
    \centering
    \includegraphics[width=1\linewidth]{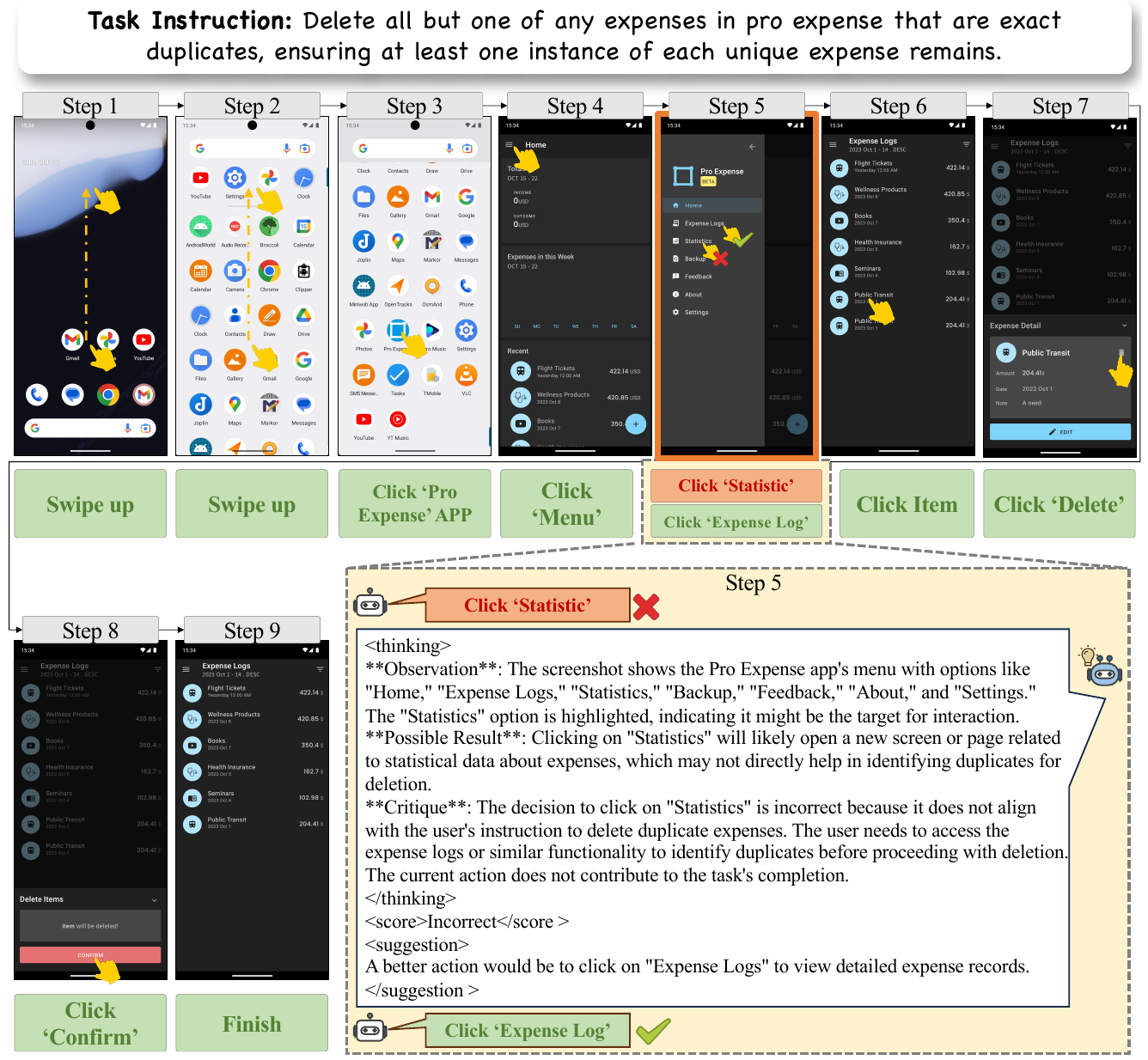}
    \vspace{-2mm}
    \caption{An instance of GUI-Critic-R1 correcting an erroneous decision within GUI automation through a pre-operative critique approach.}
    \label{fig:more_case1}
\end{figure*}

\subsection{Additional Case Study}

Figure~\ref{fig:more_case1} illustrates an operational process of GUI automation with pre-operative critic by our GUI-Critic-R1. 
In this example, the GUI agent initially demonstrates correct behavior by opening the Pro Expense app and accessing the menu. 
However, at step 5, the agent erroneously decides to click the ``statistics" button, which is incorrect. 
Our GUI-Critic-R1 model successfully identifies this as an incorrect action and analyzes the reason behind the error. 
The model determines that selecting this button would navigate to a new interface displaying expense statistics, which is irrelevant to the instruction requiring the removal of duplicate expenses. Furthermore, our model provides a suggestion for correction: click on ``Expense Logs" to view detailed expense records.

\section{GUI-Critic Dataset}
To construct the GUI-Critic-Train and GUI-Critic-Test datasets, we extract step-level data from publicly available GUI operation datasets, encompassing user instructions, current interface screenshots, text-based operative history, and corresponding correct actions. In the following section, we provide an overview of the datasets employed in our study. 

\subsection{Public Datasets}
Firstly, we review the publicly available datasets employed in our research.

\textbf{Android in the Wild (AITW)}~\cite{rawles2023androidinthewild} dataset comprises human demonstrations of device interactions, encompassing both screen captures and actions, alongside corresponding natural language instructions. It includes a substantial collection of 715,000 episodes covering 30,000 unique instructions, across four versions of the Android operating system (v10–13) and eight different device models ranging from the Pixel 2 XL to Pixel 6, each with distinct screen resolutions. The dataset involves complex multi-step tasks necessitating nuanced semantic understanding of both language and visual context.

\textbf{Android-In-The-Zoo (AITZ)}~\cite{zhang2024android} dataset contains 18,643 screen-action pairs together with chain-of-action-thought annotations, spanning over 70 Android apps, coupled with 4× useful annotations compared with action coordinate labels only. 
Based on the screen episodes from AITW~\cite{rawles2023androidinthewild}, the authors generate candidate answers for the screen descriptions, action thinkings and next action descriptions. 
These candidates are further validated and refined by human to guarantee alignment with the screenshots.

\textbf{GUI Odyssey}~\cite{lu2024gui} is a comprehensive dataset for training and evaluating cross-app navigation agents. 
GUI Odyssey comprises 7735 episodes, meticulously curated from 6 different mobile devices such as Pixel Pro and Tablet.  It encompasses 6 types of cross-app navigation tasks spanning from general system tool use to media entertainment, requiring navigating through 201 different apps and 1399 app combos from various fields such as video, music, reading

\textbf{Android Multi-annotation EXpo (AMEX)}~\cite{chai2024amex} is a comprehensive, large-scale dataset designed for generalist mobile GUI-control agents.  
Their capabilities of completing complex tasks by directly interacting with the graphical user interface (GUI) on mobile devices are trained and evaluated with the proposed dataset. 
AMEX comprises over 104K high-resolution screenshots from 110 popular mobile applications, which are annotated at multiple levels.  
AMEX includes three levels of annotations: GUI interactive element grounding, GUI screen and element functionality descriptions, and complex natural language instructions, each averaging 13 steps with stepwise GUI-action chains.

\textbf{GUICourse}~\cite{chen2024guicourse}. In this paper, the author constructed GUIAct, a GUI navigation dataset in website and android scenarios for enhancing VLMs’ knowledge of GUI systems, including 67k single-step and 15k multi-step action instructions. 

\subsection{GUI-Critic Dataset Components}

\begin{figure}[t]
    \centering
    \includegraphics[width=1.0\linewidth]{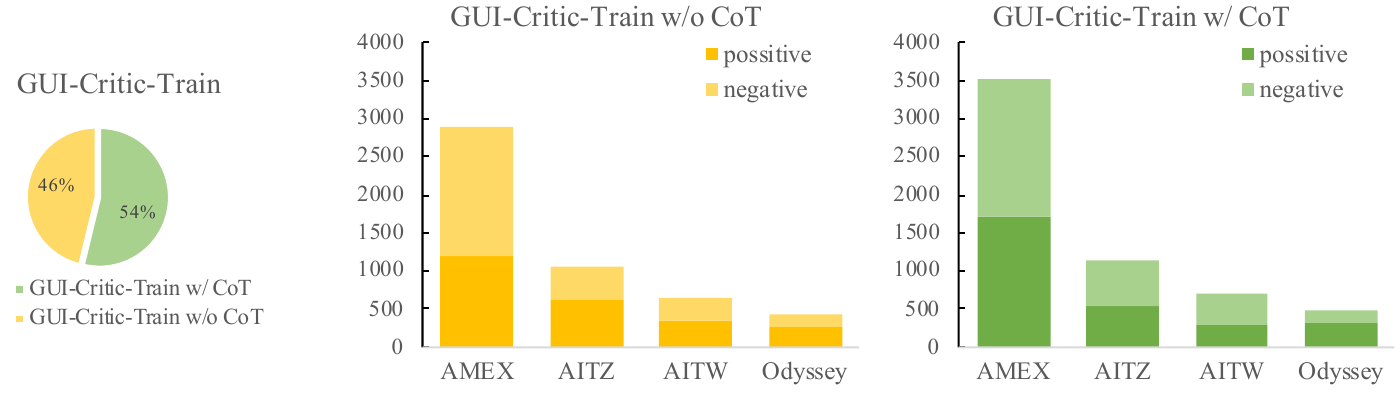}
    \vspace{-2mm}
    \caption{Illustration of the data composition for the GUI-Critic-Train dataset. The left displays the proportion of data without Chain-of-Thought (CoT) annotations versus with CoT.
    The central part shows the sources of data without CoT annotations, along with the ratio of correct to incorrect actions. 
    The right section depicts the sources of data with CoT annotations, also detailing the ratio of correct to incorrect actions.}
    \vspace{-2mm}
    \label{fig:dataset}
\end{figure}

In this subsection, we delineate the origins, constituent parts, and foundational principles of the construction of our GUI-Critic-Train and GUI-Critic-Test datasets.

\begin{table*}[h]
	\centering
	\renewcommand{\arraystretch}{0.85}
	\setlength{\tabcolsep}{8pt}
	\scalebox{0.95}{
	\begin{tabular}{p{15cm}}
        \hline
		\toprule
  \textbf{System}\\ \midrule
  There is a multimodal agent that can perform a series of actions on a smart device (phone or PC) to automate the completion of user instructions. \\
Possible actions include "click" / "left\_click" at (x,y) position, "long press" at (x,y) position, "swipe" from (x1,y1) to (x2,y2), "scroll" down or up, "drag" from (x1,y1) to (x2,y2), "type" (text content), "back", "home", "enter" and so on. \\

User instructions are usually complex and may include several detailed requirements. In some steps, the action decided by the mobile agent may be wrong. \\
Now, you are a critic model used to evaluate the agent's decision. I will provide you with the following information: \\
1. User instruction. \\
2. History: The action history of the agent in the previous steps. \\
3. Decision: The decision of the agent for this step. \\
4. Image: The screenshot before executing this action. If the action contains positional parameters (such as click and swipe), the interaction area is marked with a translucent red circle or red arrow. \\
Firstly, you need to understand the purpose of the decision. Pay attention to analyzing the interface elements in the screenshot (such as button position, text content, etc.). If there are red marks, focus on the action position. You can take appropriate account of the history information. \\
Then, based on the given information, carefully analyze the decision given by the agent for the current step: \\
1. Decision Analysis \\
(1). Observation: Observe the screenshot and analyze the state without considering the user's instruction. \\
- Focus on the operable or informative elements related to the operational decision. \\
(2). Possible Result: Speculate the most possible result of executing this decision. \\
- Predicts the screenshot change after the operation. \\
- Whether to promote the progress of core tasks. \\
(3). Critique: Determine whether the decision is correct and explain why. \\
- Focus on historical operations. \\
- Based on the previous analysis and the history, determine if this decision supports the completion of the instruction. \\
- Only perform actions specified in the instructions. \\
- Home button is the correct choice for switching apps. \\
- Both clicking a suggestion and Enter are correct when searching. \\

2. Based on the above analysis, determine whether this decision is "Correct" or "Incorrect". \\
3. Reflection: If correct, retell the action; if incorrect, suggest a better action. Propose a one-step action for the current observation, like click, swipe (with direction), type (with information), Home, Back, or Terminate (in 20 words). \\
\\
Assess the current decision's correctness in the following format: \\
<thinking> \\
**Observation**: Describe the screenshot. \\
**Possible Result**: Analysis from the possible result perspective. \\
**Critique**: Criticize why the decision is correct or incorrect. \\
</thinking> \\
<score> \\
Correct or Incorrect \\
</score> \\
<suggestion> \\
If correct, provide a brief summary; if incorrect, suggest a better decision briefly. \\
</suggestion> \\
  \midrule
  \textbf{User}\\ \midrule
       Below is the information for the current step: \\
1. User instruction: 
\{User's instruction\} \\
2. History: 
\{Operation History\} \\
3. Decision: 
\{Action\} \\
4. Image is the screenshot of this step. <image> \\
        \bottomrule
        \hline
	\end{tabular}
	}
    \caption{The prompt for the pre-operative critic.}
\label{tab:pre_critic}
\end{table*}

\begin{table*}[t]
	\centering
	\renewcommand{\arraystretch}{0.85}
	\setlength{\tabcolsep}{8pt}
	\scalebox{0.95}{
	\begin{tabular}{p{15cm}}
        \hline
		\toprule
  \textbf{System}\\ \midrule
There is a multimodal agent that can perform a series of actions on a smart device (phone or PC) to automate the completion of user instructions. \\
Possible actions include "click" / "left\_click" at (x,y) position, "long press" at (x,y) position, "swipe" from (x1,y1) to (x2,y2), "scroll" down or up, "drag" from (x1,y1) to (x2,y2), "type" (text content), "back", "home", "enter" and so on. \\

User instructions are usually complex and may include several detailed requirements. In some steps, the action decided by the mobile agent may be wrong. \\
Now, you are a critic model used to evaluate the agent's decision. I will provide you with the following information: \\
1. User instruction. \\
2. History: The action history of the agent in the previous steps. \\
3. Decision: The decision of the agent for this step. \\
4. Images: The screenshots before and after executing this action.  \\

Firstly, you need to understand the purpose of the decision. Pay attention to analyzing the interface elements in the screenshot (such as button position, text content, etc.). If there are red marks, focus on the action position. You can take appropriate account of the history information. \\
Then, based on the given information, carefully analyze the decision given by the agent for the current step: \\
1. Decision Analysis \\
(1). Observation: Observe the screenshots and analyze the state without considering the user's instruction. \\
- Focus on the operable or informative elements related to the operational decision. \\
(2). Critique: Determine whether the decision is correct and explain why. \\
- Focus on historical operations. \\
- Ensure compliance with each specific requirement in the instructions. \\
- Only perform actions specified in the instructions. \\
- Home button is the correct choice for switching apps. \\
- Both clicking a suggestion and Enter are correct when searching. \\

2. Based on the above analysis, determine whether this decision is "Correct" or "Incorrect". \\
3. Reflection: If correct, retell the action; if incorrect, suggest a remedial action on the screen after executing this action. Propose a one-step action for the current observation, like click, swipe (with direction), type (with information), Home, Back, or Terminate (in 20 words). \\
\\
Assess the current decision's correctness in the following format: \\
<thinking> \\
**Observation**: Describe the screenshots. \\
**Critique**: Criticize why the decision is correct or incorrect. \\
</thinking> \\
<score> \\
Correct or Incorrect \\
</score> \\
<suggestion> \\
If correct, provide a brief summary; if incorrect, suggest a remedial decision briefly. \\
</suggestion> \\

  \midrule
  \textbf{User}\\ \midrule
Below is the information for the current step: \\
1. User instruction: 
\{User's instruction\} \\
2. History: 
\{Operation History\} \\
3. Decision: 
\{Action\} \\
4. Images are screenshots before and after executing this action.  \\
        \bottomrule
        \hline
	\end{tabular}
	}
    \vspace{-2mm}
    \caption{The prompt for the post-operative critic.}
\label{tab:post_critic}
\end{table*}

\begin{table*}[t]
	\centering
	\renewcommand{\arraystretch}{0.85}
	\setlength{\tabcolsep}{8pt}
	\scalebox{0.9}{
	\begin{tabular}{p{15cm}}
        \toprule
		The following two sentences describe the operation that a mobile agent needs to perform on mobile in order to complete a certain user instruction: \\
		1. \{Annotated Suggestion\}\\
		2. \{Generated Suggestion\}\\
        Determine whether these two sentences describe a similar action? If yes, answer 1, if not 0, no explanation required. \\
        \bottomrule
	\end{tabular}}
    \vspace{-2mm}
    \caption{The prompt for evaluating the similarity between two action-descriptive sentences.}
\label{tab:action_similarity}
\end{table*}

\begin{table*}[t]
	\centering
	\renewcommand{\arraystretch}{0.85}
	\setlength{\tabcolsep}{8pt}
	\scalebox{0.9}{
	\begin{tabular}{p{15cm}}
        \toprule
        Note that your last decision may be incorrect! Please make a remedial decision based on the following Critiques, and consider adhering to the Suggestion provided: \{Critique\} \\
        \bottomrule
	\end{tabular}}
    \vspace{-2mm}
    \caption{The prompt for integrating critique into the GUI decision-making process.}
\label{tab:gui_agent}
\end{table*}

\subsubsection{GUI-Critic-Train}
\begin{figure}[t]
    \centering
    \includegraphics[width=1.0\linewidth]{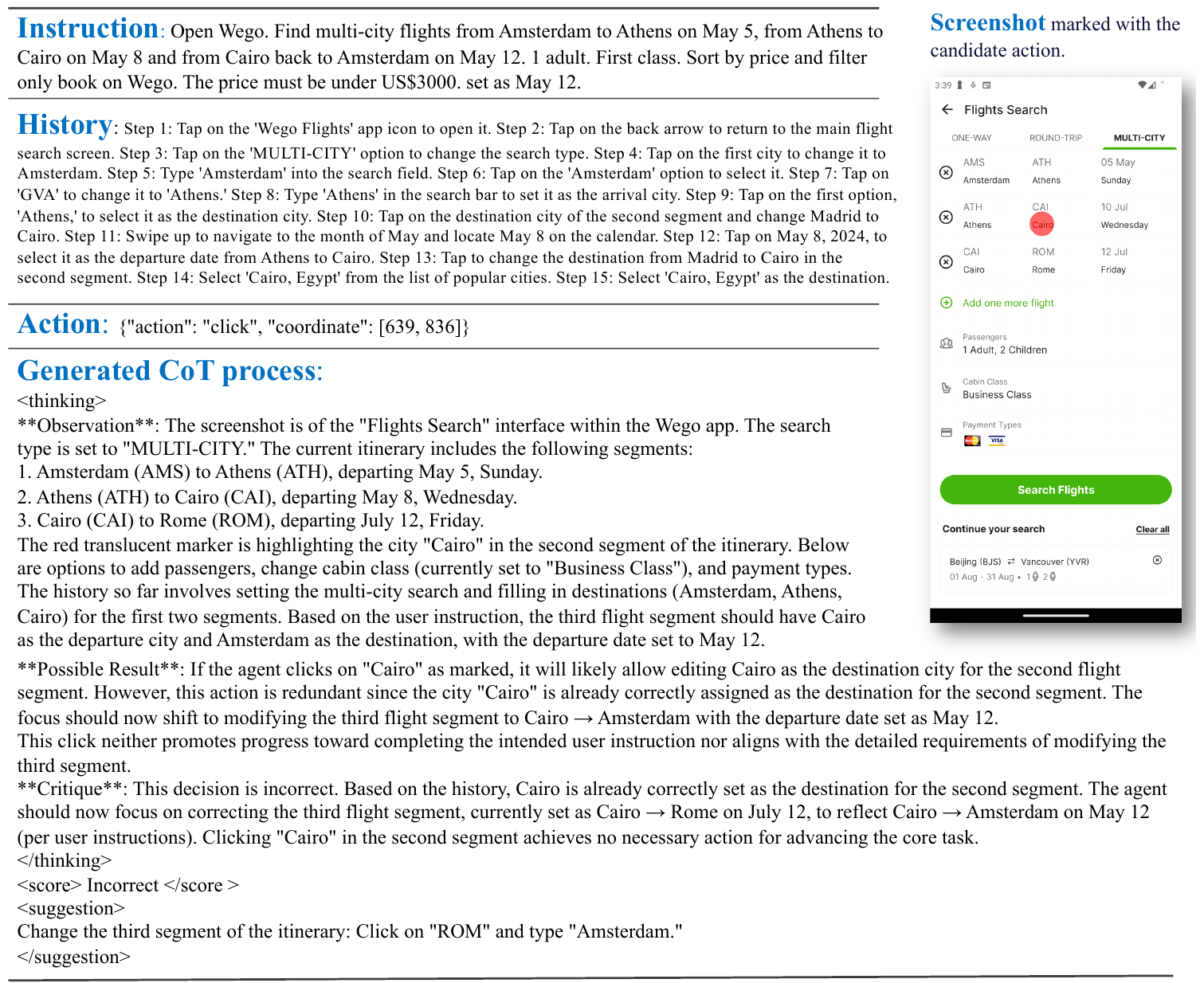}
    \vspace{-5mm}
    \caption{A data sample in GUI-Critic-Train dataset.}
    \label{fig:data_sample}
\end{figure}

Figure~\ref{fig:dataset} provides a visualization of the data sources utilized in our training set. 
For the construction of our training dataset, we predominantly selected data from AMEX\cite{chai2024amex} due to its extensive repository of high-quality GUI operation data.  
The raw data encompasses decision-making processes enriched with reasoning, which contributes to the reliability of the annotated correct actions.
Detailed methodologies for the generation of incorrect decisions and the formulation of Chain-of-Thought (CoT) critiques are elaborated in the main text (Section 3.2).
We preserved a higher proportion of samples containing incorrect decisions to facilitate the model’s acquisition of enhanced error correction competencies. 
Figure~\ref{fig:data_sample} illustrates a data sample with the generated CoT process in GUI-Critic-Train dataset. 

\subsubsection{GUI-Critic-Test}

\textbf{Mobile-Instruction Generalization (GUI-I). }
Data in GUI-I originates from the AMEX dataset~\cite{chai2024amex}. We ensure that the instructions in GUI-I are distinct from those in GUI-Critic-Train. 
After human annotation, GUI-I retains 656 GUI screenshots, each accompanied by a user instruction, operational history, and a set of candidate actions.

\textbf{Mobile-Scenario Generalization (GUI-S). }
We collect 114 test data from the Odyssey dataset~\cite{lu2024gui} for GUI-I. 
We select the data with applications different from those in GUI-Critic-Train. 
Specifically, the novel applications are: 

\begin{enumerate}
    \item \textit{TikTok}: A popular social media application for sharing and watching short-form videos, fostering a global community of creators and viewers.
    \item \textit{Chaton}: An innovative application powered by intelligent conversation models, designed to enhance user interactions through AI-driven dialogues.
    \item \textit{ClevCalc}: A multifunctional calculator tool app that offers a wide range of mathematical operations and unit conversions for everyday use.
    \item \textit{Remix}: A versatile audio editing application that allows users to mix, cut, and enhance soundtracks with a variety of editing features.
    \item \textit{Redfin Houses}: A comprehensive app for real estate transactions, providing listings, pricing, and market insights for prospective home buyers and sellers.
    \item \textit{Tripadvisor}: A travel review platform app where users can rate and explore travel destinations, accommodations, and restaurants worldwide.
\end{enumerate}

\textbf{Web-Scenario Generalization (GUI-W). } It contains 418 samples of web automation, which are randomly sampled from the GUICourse~\cite{chen2024guicourse}. 
Although the action space in web environments differs from mobile platforms (e.g., limitations on swipe direction and the incorporation of double-tapping), the underlying operational logic largely remains consistent across both scenarios.

\section{Agent Prompt}
In Table~\ref{tab:pre_critic}, Table~\ref{tab:post_critic}, Table~\ref{tab:action_similarity}, and Table~\ref{tab:gui_agent}, we present the prompts for pre-critic, post-critic, action similarity evaluation, and adding critiques to the GUI agent decision process. 
Note that, we conduct minimal modifications to the pre-critic prompt, resulting in the post-critic prompt. 
These modifications include the removal of predictions concerning possible results and the adjustment of requirements from generating corrective suggestions to remedial suggestions in the \texttt{<suggestion>...</suggestion>}. 
The action similarity evaluation prompt is leveraged in calculating the suggestion reward in S-GRPO and the suggestion accuracy.